\documentclass[letterpaper, 10 pt, conference]{ieeeconf}  

\IEEEoverridecommandlockouts                              
\usepackage[pdftex]{graphicx}
\usepackage{tabularx}
\usepackage{multirow}

\usepackage{amsmath}
\usepackage{amssymb}
\usepackage{amsfonts}
\usepackage[ruled, vlined, linesnumbered]{algorithm2e}

\usepackage{balance}

\usepackage{graphicx, subcaption}

\usepackage{url}

\usepackage{xcolor}

\usepackage{cite}

\usepackage{colortbl}
\definecolor{color1}{rgb}{1,0.67,0.25}  
\definecolor{color2}{rgb}{0,0.49,0.56}  
\definecolor{color3}{rgb}{0.54,0.80,0.26}  

\usepackage[belowskip=-15pt, aboveskip=4pt, font=small]{caption}
\usepackage{hyperref}

\usepackage{balance}

\title{\LARGE \bf Autonomous Exploration of Unknown 3D Environments Using a Frontier-Based Collector Strategy}

\author{Iv\'an D. Changoluisa Caiza$^{1}$, 
        Ana Milas$^{1}$, 
        Marco A. Montes Grova$^{2}$, Francisco Javier Pérez-Grau$^{2}$,
        Tamara Petrovic$^{1}$
\thanks{$^{1}$Iv\'an David Changoluisa Caiza, Ana Milas and Tamara Petrovic are with the University of Zagreb, Faculty of Electrical Engineering  and Computing, LARICS Laboratory for Robotics and Intelligent Control Systems, Unska 3, 10000 Zagreb, Croatia; {\tt\small (ivan.changoluisa, ana.milas, tamara.petrovic)@fer.hr}
\newline 
$^{2}$Francisco Javier Pérez-Grau and Marco A. Montes Grova are with the Advanced Center for Aerospace Technologies (CATEC), Aerospace Technology Park of Andalusia, Seville, Spain {\tt\small (fjperez, mmontes)@catec.aero}}}

\begin{document}

\maketitle

\begin{abstract}
Autonomous exploration using unmanned aerial vehicles (UAVs) is essential for various tasks such as building inspections, rescue operations, deliveries, and warehousing. However, there are two main limitations to previous approaches: they may not be able to provide a complete map of the environment and assume that the map built during exploration is accurate enough for safe navigation, which is usually not the case.  To address these limitations, a novel exploration method is proposed that combines frontier-based exploration with a collector strategy that achieves global exploration and complete map creation. In each iteration, the collector strategy stores and validates frontiers detected during exploration and selects the next best frontier to navigate to. The collector strategy ensures global exploration by balancing the exploitation of a known map with the exploration of unknown areas.
In addition, the online path replanning ensures safe navigation through the map created during motion. The performance of the proposed method is verified by exploring 3D simulation environments in comparison with the state-of-the-art methods. Finally, the proposed approach is validated in a real-world experiment. 
\end{abstract}

\IEEEpeerreviewmaketitle

\section{Introduction}
\label{sec:introduction}

In recent years, autonomous robots have been utilized to inspect structures, monitor and map unreachable areas and explore cluttered and large environments \cite{CamposMacas2020}, \cite{Kwon2020}, to name a few. In particular, because of the agile movement and high degree of autonomy, exploration in unknown environments using unmanned aerial vehicles (UAVs) has been widely researched \cite{Yamauchi1997, Bircher2016, Cieslewski2017, Witting2018, Dai2020, Selin2019, Batinovic-RAL-2021, Batinovic-RAL-2022}. 
Existing exploration methods using UAVs mainly rely on random sampling-based path planning, such as Rapidly-exploring Random Trees (RRT) \cite{RRT1}, which yields unnecessary movements that affect the speed of exploration  \cite{Witting2018}, \cite{Selin2019}. On the other hand, the frontier-based exploration strategies  \cite{Yamauchi1997} ensure global exploration, but demand higher computational resources. However, both strategies may result in unmapped parts of the environment, especially when exploring large environments \cite{Selin2019}, \cite{Batinovic-RAL-2021}. In the context of warehousing, some tasks require a complete and detailed map of the environment, such as inventory management, monitoring stockpiles, or inspecting hard-to-reach areas. In other words, for some tasks, it is inevitable to take into account the trade-off between the time required for exploration and the quality of the map.
Additionally, to be able to explore real-world environments and ensure safe navigation, the exploration planner should consider possible position drift and unreachability of waypoints.

To this end, this paper proposes a novel exploration method that combines frontier-based exploration with a collector strategy. The collector strategy utilizes a two-phased frontier filter that combines an obstruction checker and a proximity checker to speed up the exploration. Clusters of frontiers are saved during the exploration process to balance the exploitation of the known area and the exploration of an unknown area. The algorithm plans the path to the next best frontier and checks the validity of the path during execution. If necessary, the path is replanned to ensure safe and obstacle-free navigation. 
\begin{figure}[t]
  \centering
  \begin{minipage}{0.5\columnwidth}
    \centering
    \includegraphics[width=\columnwidth]{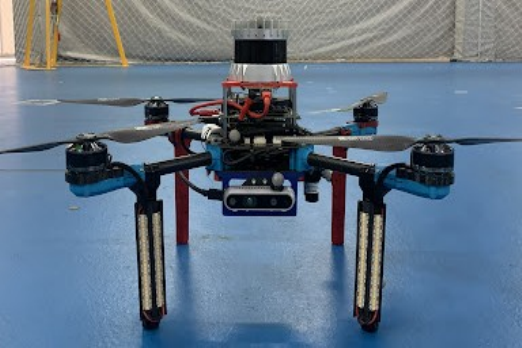}
    \subcaption{}
    \label{fig:uav}
  \end{minipage}
  \hfill
  \begin{minipage}{0.485\columnwidth}
    \centering
    \includegraphics[width=\columnwidth]{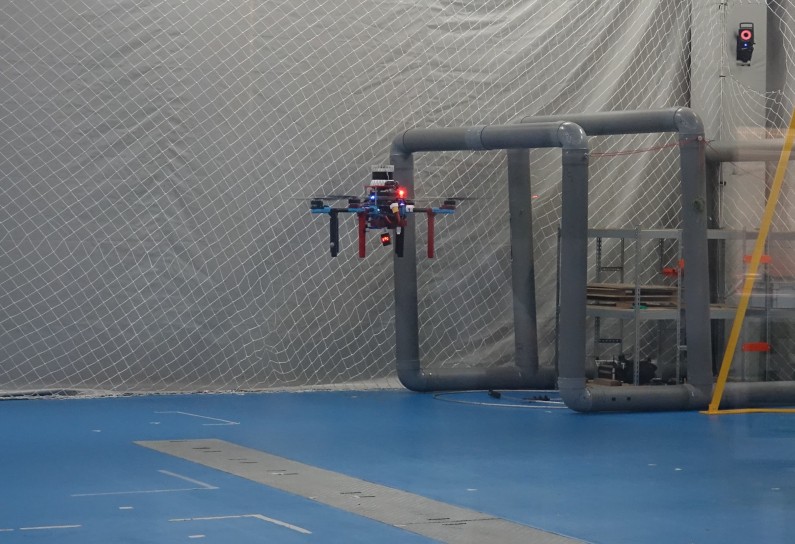}
    \subcaption{}
    \label{fig:flying_area}
  \end{minipage}
  \vspace{12pt}
  \caption{Exploration in a real-world environment. (a) A \textit{HOLYBRO X500} quadcopter equipped with an \textit{Ouster 128-U} LiDAR. (b) The UAV performs autonomous exploration,  running the algorithm onboard.}
  \label{fig:real_world}
  \vspace{-5 pt}
\end{figure}
The performance of the proposed approach is validated by an experimental comparison with the state-of-the-art exploration methods in simulation environments. Furthermore, a real-world experiment is conducted to demonstrate the capability of our approach.
The main contributions of this work are summarized as follows:
\begin{itemize}    
    \item A collector strategy that stores, filters and selects the best frontier during exploration. It ensures complete map acquisition and balances the exploitation of known map data and the exploration of unknown areas.
    \item Online replanning algorithm that allows continuous checking of path validity and replanning when needed. It ensures the execution of collision-free paths and thus safe navigation. 
    \item Evaluation and demonstration of the proposed approach in both simulation and a real-world experiment.
\end{itemize}
Everything presented in this paper is released as open source \cite{github-collector} to benefit the community, enable replication of experiments and easy integration.

The rest of this paper is structured such that Section \ref{sec:related} summarizes the related work. Section \ref{sec:proposed} defines the problem and describes the proposed approach in detail. Evaluation during simulation and validation in a real-world experiment are discussed in Sections \ref{sec:simulation} and \ref{sec:experiment}, respectively. Finally, the conclusions will be addressed in Section \ref{sec:conclusion}.





\section{Related work}
\label{sec:related}

Numerous studies have addressed the autonomous exploration using UAVs, with most approaches falling into three categories: frontier-based, sampling-based, and hybrid strategies.  

Frontier-based approaches explore the environment by targeting areas on the frontier between the explored and unexplored environments. This idea was first introduced by Yamauchi in \cite{Yamauchi1997}, where the next best goal is the closest frontier. Similarly, in \cite{Cieslewski2017}, the next best goal is the frontier that minimizes the velocity change to maintain a consistently high flight speed. Additionally, \cite{Batinovic-RAL-2021} addresses the high computational cost associated with frontier exploration using LiDAR scanners by utilizing the multi-resolution capabilities of the OctoMap data structure to detect frontiers at a coarse level.
Frontier-based exploration approaches for 3D environments are also researched in \cite{Zhu2015, Dai2020, Faria2019}.

While frontier-based methods are suitable for large environments due to their ability to identify unexplored spaces in the global map, sampling-based approaches are well-suited for cluttered spaces but can suffer from local minima issues, hindering complete coverage of the target environment \cite{zhou2020survey}.
Bircher et al. \cite{Bircher2016} introduced sampling-based techniques in UAV exploration. 
Building upon this approach, Schmid et al. \cite{Schmid2020} introduced an informative path planning algorithm inspired by Rapidly-exploring Random Tree Star (RRT*).
To accelerate information gain computation, \cite{Batinovic-RAL-2022} proposed a novel technique based on the recursive shadowcasting algorithm. 
Hybrid strategies combine the advantages of both frontier-based and sampling-based approaches as shown in \cite{Selin2019}, \cite{Respall2021}, \cite{zhou2021fuel} and \cite{zhou2023racer}.

Regarding online replanning during autonomous exploration, the authors in \cite{Oleynikova2018} propose a local exploration method for replanning in cluttered environments for UAVs to enable fast reactions to newly observed parts of the environment. Furthermore, Zhao et al. \cite{Zhao2023} introduce the method based on the framework presented in \cite{zhou2021fuel} and adapting dynamic replanning strategy to avoid the stop-and-go maneuvers. 

Efforts to improve the efficiency, safety, accuracy, and robustness of autonomous exploration have shown promising results. However, to the best of our knowledge, no frontier-based exploration algorithm for UAV onboard applications has yet been developed that focuses on balancing the use of known map data with the exploration of unknown areas. Additionally, state-of-the-art methods focus on replanning to speed up exploration and avoid obstacles instead of dealing with replanning due to uncertainties in the map. 
With this in mind, we present a novel autonomous exploration strategy specifically designed for UAVs with limited payload capabilities and computational resources. Our approach integrates real-time mapping, exploration and navigation capabilities directly onboard the UAV.


\section{Proposed approach}
\label{sec:proposed}

The main goal of the proposed approach is to explore a bounded and previously unknown 3D space $V \subset \mathbb{R}^{3}$.
As a basis for our approach, an OctoMap $M$ is used, a hierarchical volumetric 3D representation of the environment \cite{Hornung2013}. Each cube of the OctoMap, with a predefined resolution $r$, is denoted as a voxel (cell) $\mathbf{v}$, which can be \textit{free}, \textit{occupied} or \textit{unknown}. Free voxels form the free space $V_{free} \subset V$, occupied voxels form the occupied space $V_{occ} \subset V$ and unknown voxels form the unknown space $V_{un} \subset V$. Initially, the entire bounded space is unknown, $V \equiv V_{un}$, and the unknown space decreases as the exploration advances. The entire space is a union of the three subspaces $V \equiv V_{free} \cup V_{occ} \cup V_{un}$.
The exploration problem is considered fully solved when $V_{occ} \cup V_{free} \equiv V \setminus V_{res}$, where $V_{res}$ is residual space defined as an unexplored space, which remains inaccessible to the sensors. 

The UAV is represented with a state vector $\mathbf{x} = \begin{bmatrix} \mathbf{p}^T & \psi \end{bmatrix}^T \in \mathbb{R}^4$ that consists of the position $\mathbf{p} = \begin{bmatrix} x & y & z \end{bmatrix}^T \in \mathbb{R}^{3}$ and the yaw rotation angle around the body $z$ axis $\psi \in [-\pi, \pi)$. For collision checking, it is considered that the UAV is inside a rectangular prism centered at $\mathbf{p}$, with adequate length, width and height $l$, $w$, $h$.
A sensor is defined with a maximum range $R_{max} \in \mathbb{R}$ with horizontal and vertical Field of View (FOV) in range, $\alpha_{h}$, $\alpha_{v}$ $\in (0^{\circ}, 360^{\circ}]$, respectively.

The previously developed frontier-based method \cite{Batinovic-RAL-2021} is used to extract frontier voxels (frontiers) from the OctoMap. The frontiers are clustered and a collector strategy is utilized to select the next best goal. Finally, the UAV plans a path to the next best frontier and navigates while simultaneously checking the validity of the path. The main contributions in this part are the proposed collector strategy and the online replanning algorithm used for obstacle-free navigation.

\subsection{UAV Exploration Framework}
\label{subsec:framework}

\begin{figure}[t!]
  \centering
  \includegraphics[width=1\columnwidth]{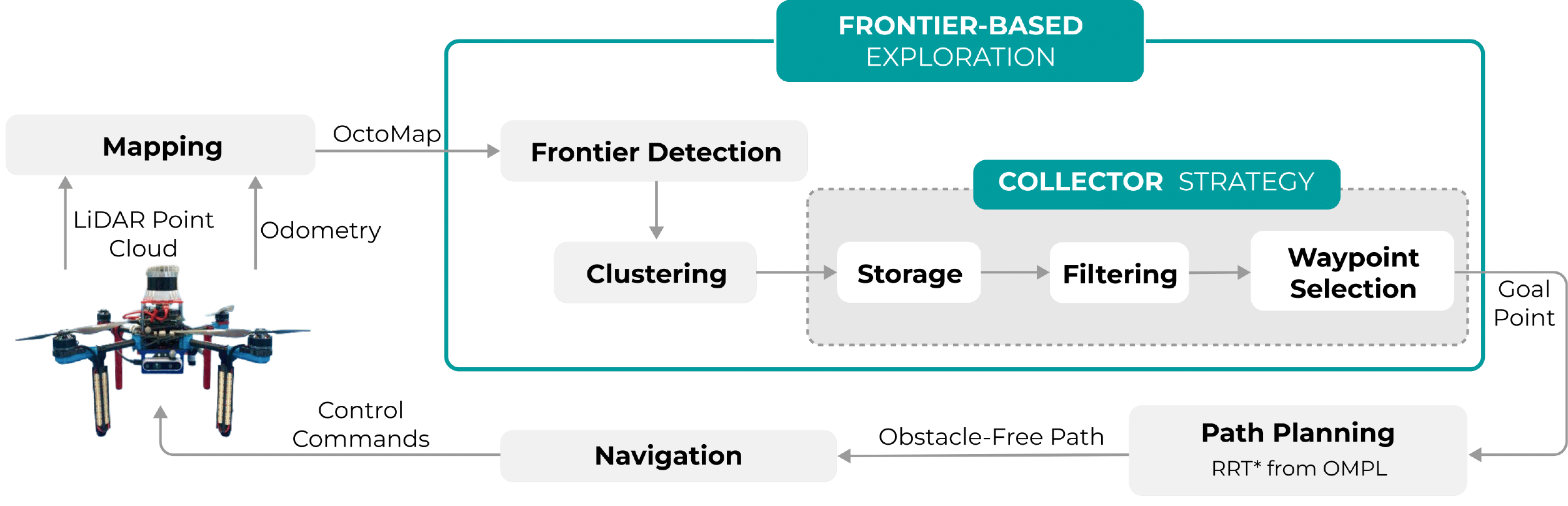}
  \caption{Overall schematic diagram of the 3D exploration. The LiDAR point cloud and odometry data represent inputs to the mapping module that creates an OctoMap. The frontier-based exploration module (highlighted in green) includes our proposed collector strategy, which generates a goal point that is passed to the path planning module. Finally, the robot navigates to the goal point by executing an obstacle-free path.}
  \label{fig:exp_frame}
  \vspace{-5pt}
\end{figure}

The proposed framework for autonomous exploration and safe navigation encompasses a few modules. As shown in Fig. \ref{fig:exp_frame}, it contains the mapping module that systematically stores the gathered environmental information, enabling the definition of obstacle locations, shapes, and free spaces, thereby determining the navigable areas for the UAV. The exploration module utilizes the OctoMap to determine the next goal point and passes it to the path planning module. Once the path is planned, the navigation module generates control commands to guide the UAV along the path. 

An OctoMap is generated using the LiDAR data and is used for both frontier detection and collision-free navigation. Note that our exploration strategy is applicable to various types of autonomous robots equipped with LiDARs or other sensors that can be used to build an OctoMap.
The exploration process is iterative and is repeated until there are no more frontiers to be explored. The OctoMap is continuously updated as the UAV moves toward the goal. Once the goal is reached, the next planning iteration begins. The integration of these interconnected systems ensures efficient and comprehensive exploration and mapping of the environment by the UAV.

\subsection{Frontier Detection and Clustering}
\label{subsec:frontier}

A frontier, $F$, can be defined as a set of voxels $\mathbf{v}_f$ with the following property \cite{Batinovic-RAL-2021}:

\begin{equation}
    F = \{\mathbf{v}_f \in V_{free} : \exists neighbor(\mathbf{v}_f) \in V_{un}\}. 
    \label{eq:front_def}
\end{equation}
In other words, a frontier consists of free voxels with at least one unknown neighbor. The center of a frontier voxel is often called the frontier point. 

The OctoMap $M$ used for frontier detection is generated using LiDAR scans $S$.
The OctoMap is in the form of octrees, a format suitable for path planning. During the exploration, the OctoMap $M$ is built iteratively using the method described in\cite{Hornung2013}. 
The current OctoMap $M^i$ is created from the current LiDAR scan $S^i$ added to the OctoMap explored so far.
At the same time, a frontier detection cycle is performed periodically to ensure that frontiers are constantly updated. Note that the rate of an OctoMap update process is lower than the frontier detection process since the OctoMap update is a computationally demanding process, especially when using dense scans.

Let $V^i_{free}$ and $V^{i-1}_{free}$ correspond to the free voxels in two consecutive OctoMaps, $M^i$ and $M^{i-1}$.  
Then the local frontier $F_l$ contains only newly created frontier points \cite{Batinovic-RAL-2021}:
\begin{equation}
F_l = \{\mathbf{v}_f \in V^i_{free} \setminus V^{i-1}_{free}: \exists neighbor (\mathbf{v}_f) \in V^i_{un}\}.   
\end{equation}

In the proposed approach, to get frontier voxels which are candidates for exploration, denoted as $F_c$, frontiers $F_l$ are clustered using the mean shift clustering algorithm \cite{Fukunaga1975}. 
The computationally most complex component of the mean-shift procedure is the identification of the neighbors of a point in 3D space (as defined by the kernel and its bandwidth). The kernel represents a weighting function and applying it to 3D points generates a probability surface (e.g., a density function). The kernel bandwidth regulates the size of the "window" over which the mean is calculated. In this paper, the Gaussian kernel is used.
To make the mean-shift algorithm work in real time, an appropriate bandwidth is carefully selected to balance between computation time and the desired outcome with respect to the size of the environment and OctoMap resolution $r$.

\subsection{Collector Exploration Strategy}
\label{subsec:collector}
After frontiers are identified and clustered, the proposed collector strategy processes the clustered frontiers (candidates) and outputs the next best frontier to be visited. 
Each candidate frontier $\mathbf{v}_c \in F_c$ has \textit{information gain}, i.e., a measure of the unexplored region of the environment potentially visible from $\mathbf{v}_c$. 
The information gain $I(\mathbf{v}_c)$ is defined as the share of unknown voxels in a cube placed around $\mathbf{v}_c$. The size of the cube is defined with respect to the range of the sensor used. Often the information gain is estimated using a ray tracing algorithm and a real sensor FOV. In this paper, the information gain is calculated using a cube-based approximation \cite{Batinovic-RAL-2021}. By using the proposed simplification, the high calculation effort required by ray tracing is avoided.

\begin{figure}[t]
  \centering
  \includegraphics[width=1\columnwidth]{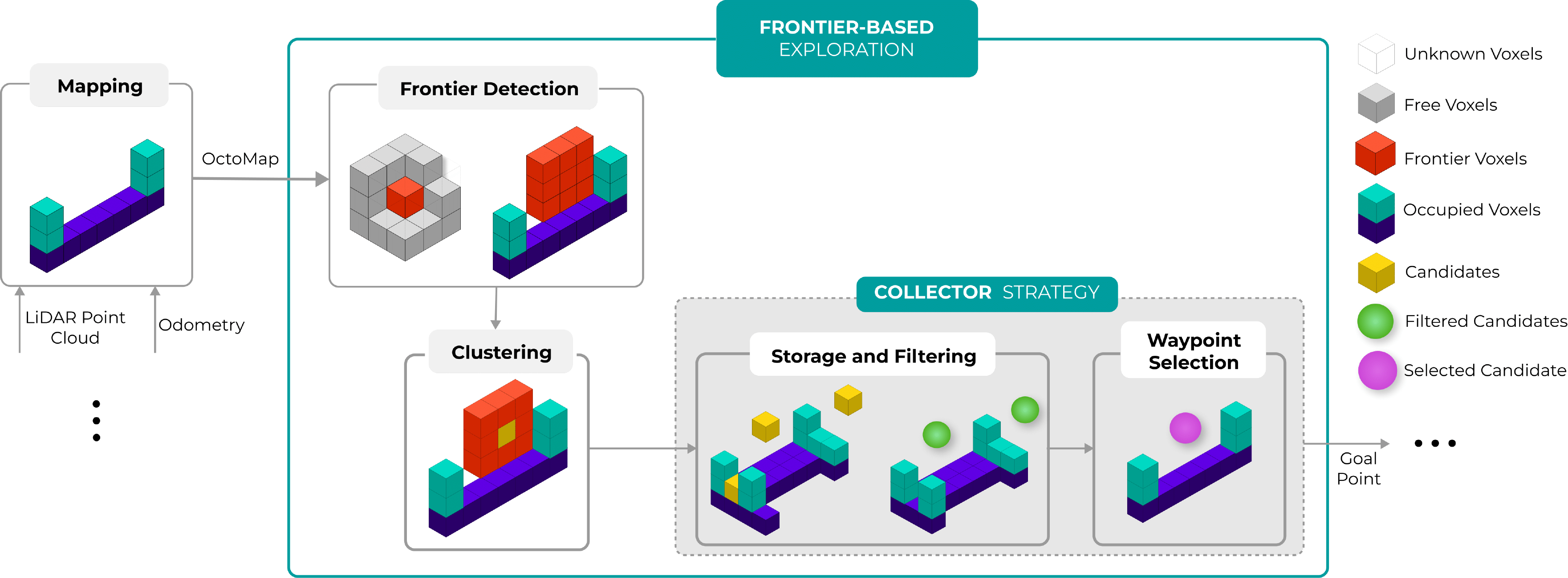}
  \caption{Detailed description of the individual modules of the proposed frontier-based exploration strategy on the small part of the OctoMap.}
  \vspace{-5pt}
  \label{fig:system_detailed}
\end{figure}

First, all candidates $F_c$ at each stage of the exploration are stored in a set of global candidates $F_{gc}$. Second, $F_{gc}$ is filtered at each iteration to eliminate obstructed and proximal points, previously visited points, or points with low information gain. In the third stage, the closest global candidate is selected as the best frontier and forwarded to the path planner. The collector strategy is shown in Fig. \ref{fig:system_detailed}.

\begin{figure}[t]
  \centering

  \begin{minipage}{0.3\columnwidth}
    \centering
    \includegraphics[width=\columnwidth]{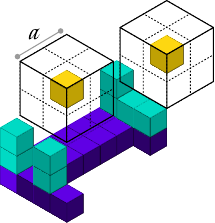}
    \subcaption{}
    \label{fig:obstructed}
  \end{minipage}
  \hspace{1cm}
  \begin{minipage}{0.37\columnwidth}
    \centering
    \includegraphics[width=\columnwidth]{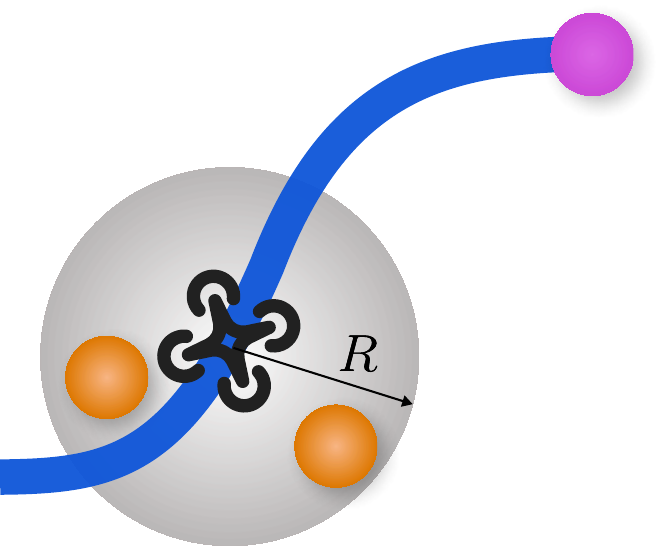}
    \subcaption{}
    \label{fig:similar}
  \end{minipage}
  \vspace{12pt}
  \caption{Implemented checkers within the collector strategy. (a) The obstruction checker is used to remove candidates that contain occupied voxels within the cube around the candidates. (b) The proximity checker eliminates candidates (orange points) inside the radius $R$ from the current UAV position while the UAV executes the planned path to the goal point (pink point).}
  \vspace{-5pt}
  \label{fig:state_checkers}
\end{figure}

Sometimes, due to the OctoMap update, the candidate may be denoted as obstructed.
An obstructed candidate is defined as a candidate $\mathbf{v}_c$ whose (at least one) neighbor is occupied. The neighbors of the candidate $\mathbf{v}_c$ are determined by setting a cube centered on the candidate position (similar to the information gain calculation), as shown in Fig. \ref{fig:state_checkers} (a). The cube size $a$ is determined in accordance with the path planner validity checker (Section \ref{subsec:path_planning}). The cube is sampled along the three axes using the predefined OctoMap resolution $r$. Should any of the nodes within the cube be identified as occupied, the candidate is considered invalid and removed from the set of global candidates $F_{gc}$. $F_{gc}$ is updated in each iteration as follows:
\begin{equation}
    F^{i}_{gc} =  F^{i}_{gc} \backslash \{\mathbf{v}_c \in F^{i}_{gc}: \exists neighbor (\mathbf{v}_c) \in V^i_{occ}\}, F^0_{gc} = \emptyset\}.
\end{equation}

Furthermore, candidates in the set $F_{gc}$ may be densely distributed depending on the chosen cluster bandwidth (described in \ref{subsec:frontier}). This may lead to the selection of similar goal points and thus to back-and-forth movements. 
To address this issue, proximity filtering is proposed. The filter removes a candidate if the Euclidean distance between the UAV position $\mathbf{p}_i$ and the position of the candidate (voxel center) $\mathbf{p_{v}}_{c}$ is less than $R$, while the UAV navigates to the goal point.  
In a 3D space, this means that each candidate inside a sphere of radius $R$, centered on the current position of the UAV $\mathbf{p}_i$, is removed from $F_{gc}$. The current $F_{gc}$ is updated accordingly:
\begin{equation}
  F^{i}_{gc} =  F^{i}_{gc} \backslash \{\mathbf{v}_c \in F^{i}_{gc}: \lVert \mathbf{p}_i - \mathbf{p_{v}}_{c}\rVert < R\}, F^0_{gc} = \emptyset\}.   
\end{equation}
Fig. \ref{fig:state_checkers} (b) represents the proximity filtering process. 


Similarly to how some candidates may be obstructed by OctoMap updates, candidates may also become useless in terms of information gain. In other words, some candidates may be located in already explored zones or in regions that provide minimal information gain. With the aim of speeding up the exploration process, information gain filtering is utilized in the collector strategy. All candidates whose information gain $I(\mathbf{v}_c)$ is below the threshold $I_{threshold}$ are removed from $F_{gc}$:
\begin{equation}
  F^{i}_{gc} =  F^{i}_{gc} \backslash \{\mathbf{v}_c \in F^{i}_{gc}:  I(\mathbf{v}_c) < I_{threshold}\}, F^0_{gc} = \emptyset\}.   
\end{equation}
Finally, the collector strategy also removes all previously visited candidates. 

The best frontier $\mathbf{v}_{bf}$ is selected from the set of filtered (validated) global frontiers $F_{gc}$. This is an iterative minimization process that identifies the candidate closest to the current UAV position.
The estimated distance is approximated by the Euclidean distance between the UAV position $\mathbf{p}_i$ and the position of the candidate $\mathbf{p_{v}}_{c}$.

\subsection{Path Planning and Navigation}
\label{subsec:path_planning}
As soon as the best frontier point is selected, it is forwarded to a path planner as a waypoint. The robot starts to follow the planned path and navigates to the best frontier point ${\mathbf{v}_{bf}}$. 
The path planning module includes the RRT* algorithm, an extension of the original RRT algorithm. 
The approach utilized within this paper has been developed in our previous work \cite{Arbanas2018,Ivanovic2022ParabolicAirdrop}, and is available online \cite{larics-motion-planning}. 
In each iteration, the planner avoids occupied voxels in the OctoMap and generates a path through the free voxels up to the best frontier point. The crucial part of the planner is the state validity checker, which evaluates the validity of configurations based on system constraints such as collision avoidance and environment-specific criteria.
The path planner takes a binary representation of the OctoMap as an input, which provides an efficient and compact description of the environment. The UAV is represented as a rectangular prism of appropriate dimensions within the state validity checker.

\begin{figure}[t]
  \centering
  \includegraphics[width=0.44\columnwidth]{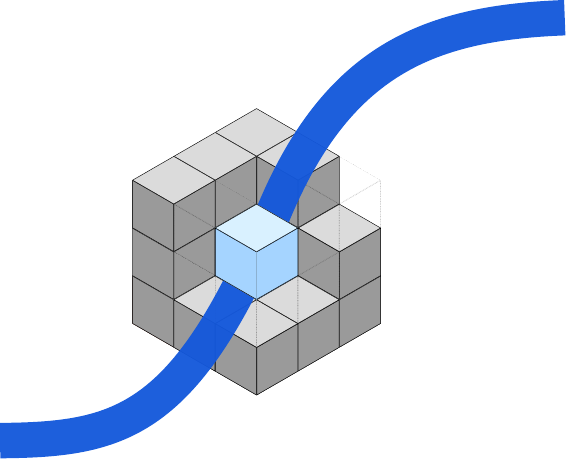}
  \caption{The validity checker for the current obstacle-free path (highlighted in blue). For each point (highlighted in light blue), the validity of the path is checked considering the set of neighbors (gray voxels) around the point.}
  \vspace{-5pt}
  \label{fig:planner_checker}
\end{figure}

During the path execution, the OctoMap is updated and newly discovered obstacles may appear in or near the path, as noticed in the experimental analysis within our previous work \cite{Batinovic-RAL-2021}. To overcome this issue, the path planner is extended to check the validity of the path during motion, as the OctoMap is updated. For each point along the path, it checks whether the UAV can execute it without interfering with obstacles. It uses a method similar to frontier detection to analyze the surroundings of a given point, as depicted in Fig. \ref{fig:planner_checker}.
If the validity checker detects an obstacle during motion, the UAV is stopped and a new path is requested, maintaining the same goal point. This ensures online replanning to the same goal point and that the UAV does not attempt to traverse newly detected obstacles.
However, if it is not possible to plan to a goal point, the goal point is discarded. 
The next best frontier point is calculated either when the previous path is discarded or after the previous frontier point is reached by the UAV. During the exploration, the number of candidates changes and once the set of valid candidates is empty, $F_{gc} = 0$, the exploration process is considered done. 

The path execution and the UAV control are achieved using an MPC-based tracking method. The original implementation is presented in \cite{tomas_mpc}. This tracking method allows the UAV to smoothly follow and quickly change the UAV trajectory based on the current system state and model dynamics. Furthermore, the tracker enables a safe and stable flight, regardless of the goal point resulting from the exploration planner.

\section{Simulation-based evaluation}
\label{sec:simulation}

Simulations are performed in the Gazebo environment using the Robot Operating System (ROS) and a quadcopter. The quadcopter is equipped with a LiDAR sensor whose maximum range is reduced to $R_{max} = 20$ m in the simulations.  It has a horizontal and vertical FOV $\alpha_{h} = 180^\circ$,  $\alpha_{v} = 30^\circ$, respectively.  For collision checking, the dimensions of a rectangular prism around the UAV are set to $l = 0.6$ m, $w = 0.6$ m, $h = 0.5$ m, while the cube size $a=1.2$ m. The maximum velocity of the UAV is set to 1 m/s, identical to the velocity used in the real-world experiment, which is low for safety reasons; however, it is comparable to the state-of-the-art (\cite{Bircher2016}, \cite{Dai2020}), especially in the experimental analysis.

The proposed exploration algorithm is compared with the greedy frontier method (GF), wherein the candidate with the highest information gain (described in Section \ref{subsec:collector}) is selected as the next best frontier. Additionally, it is compared to the more recent multi-resolution frontier planner (MRF) \cite{Batinovic-RAL-2021}. The parameters used in the MRF are set to their default values explained in \cite{Batinovic-RAL-2021}. Both the GF and the MRF are adapted to our quadcopter, equipped with a LiDAR, and to our control system to allow the fairest possible comparison. OctoMap resolution is set to $r = 0.50$ m. Additionally, results are shown as an average of 31, 37, and 8 runs of our approach, GF and MRF, respectively, with the UAV starting from 9 different positions in the environment. All simulations have been run on Intel(R)Core(TM) i7-108750H CPU @ 2.30GHz $\times$ 16.

\vspace{-5pt}
\subsection{Warehouse-like Environments}
\label{subsec:warehouse}

\begin{figure}[t]
  \centering
  \includegraphics[width=1\columnwidth]{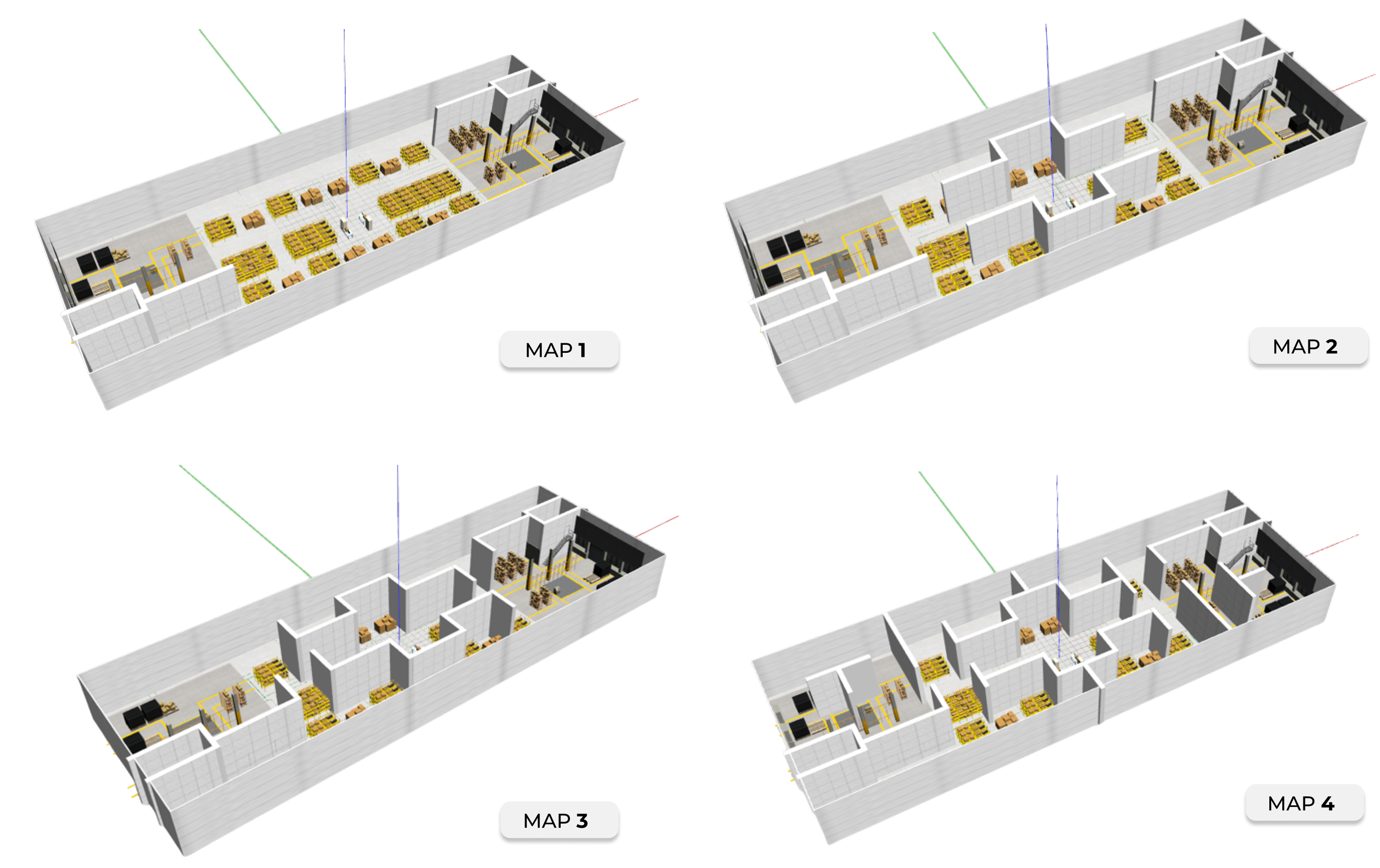}
  \caption{Warehouse-like environments used in simulation evaluation.}
  \label{fig:warehouse_maps}
  \vspace{-5pt}
\end{figure}

The approaches are compared in warehouse-like environments of varying complexity (Fig. \ref{fig:warehouse_maps}). All environments refer to 76 m $\times$ 28 m $\times$ 8 m warehouses.
This evaluation allows for a more nuanced understanding of the performance of the exploration approaches under different conditions.
  
\definecolor{Gray}{gray}{0.85}
\begin{table*}[t]
\footnotesize
\centering
\caption{Total exploration time, distance traveled and percentage of map explored for three approaches in four warehouse-like scenarios.}
  \begin{tabular}{|c|c|ccc|ccc|ccc|}
    \cline{2-11}
    
      \multicolumn{1}{c|}{} & \multirow{2}{*}{\textbf{Map}} & \multicolumn{3}{c|}{\textbf{Exploration Time [min]}} & \multicolumn{3}{c|}{\textbf{Distance Traveled [m]}} & \multicolumn{3}{c|}{\textbf{Map Explored [\%]}}\\ 
    \cline{3-11}
      \multicolumn{1}{c|}{} & & \textbf{Avg} & \textbf{Max} & \textbf{Min} & \textbf{Avg} & \textbf{Max} & \textbf{Min} & \textbf{Avg} & \textbf{Max} & \textbf{Min}\\ \hline
      
      \cellcolor{color1} & M1 & 5.38 & 6.08 & 4.38 & 237.20 & 272.42 & 195.12 & 96.85 & 100.00 & 94.35 \\ 
     \cellcolor{color1} & M2 & 8.77 & 10.13 & 6.88 &373.48  & 454.14 & 283.4 & 98.19 & 100.00 & 94.99 \\ 
    \cellcolor{color1} & M3 & 7.90 &  10.13 & 5.18 & 337.35  & 440.25 & 218.52 & 93.98 & 97.22 & 91.13\\ 
     \multirow{-4}{*}{\cellcolor{color1}{\textbf{GF}}} & M4 & 11.59 & 16.70 &7.83 & 508.23  & 686.94 & 354.08& 96.33 & 99.19 & 92.42\\ \hline

     \cellcolor{color2!80} & M1 & 6.48 & 6.67 & 6.30 & 280.17 & 288.80 & 271.4 & 96.14 & 99.32 & 92.95\\ 
     \cellcolor{color2!80} & M2 & 8.76 & 9.23 & 7.87 & 354.50 & 380.19 & 311.70 & 97.73 & 99.05 & 95.83 \\ 
     \cellcolor{color2!80} & M3 & 7.67 &10.72 & 5.30& 312.49 &390.00 &219.92 & 97.15& 100.00 &93.43 \\ 
     \multirow{-4}{*}{\cellcolor{color2!80}{\textbf{OURS}}} & M4 &9.43&10.87 &7.50 &383.00 &459.24 &318.23 &97.96 &100.00 &93.28 \\ \hline

     \cellcolor{color3} & M1 &3.45  & 3.54 & 3.36 & 162.20 & 167.8 & 156.64 & 87.92 &90.36 & 85.48\\ 
      \cellcolor{color3} & M2 &5.36 &5.42 & 5.31 & 218.60 &222.88 &214.36 & 91.76 & 94.04 & 89.47 \\ 
      \cellcolor{color3} & M3 &6.77 &9.15 &4.40 &284.38 &386.97 &181.80 & 94.74 &95.87 &93.62 \\ 
      \multirow{-4}{*}{\cellcolor{color3}{\textbf{MRF}}}  & M4 &6.28 &6.29 &6.27 &266.59 &270.87 &262.31 & 92.80 & 93.91& 91.69 \\ \hline
  \end{tabular}
  \label{tab:complete_results}
   \vspace{-15pt}
\end{table*}

The results of comparison for four different maps (from M1 to M4) are succinctly summarized in Table \ref{tab:complete_results}. The evaluation of each algorithm was performed by assessing three parameters: exploration time, distance traveled, and the proportion of the map explored. In the GF and MRF methods, exploration is considered complete when no more candidates can be found. Conversely, in our approach, the exploration concludes when there are no more valid candidates in the set of potential exploration targets.

Although the GF approach achieved exploration of close to 100\% for all maps, it showed a tendency towards longer exploration times and greater traveled distances. This was particularly evident on the complex map M4, where an average exploration time of 11.59 minutes and an average distance of 508.23 meters were recorded. This algorithm often leads to considerable back-and-forth movement, which may result in less efficient paths. 
The MRF algorithm completed the exploration in the shortest average time and covered the least distance across all maps but fell slightly in exploration completeness, especially for the more complex maps M3 and M4. In contrast, our approach showed consistent and adaptive performance across all maps, providing an efficient balance between exploration time, distance traveled, and completeness of exploration. On map M4, our approach took an average exploration time of 9.43 minutes, traveled a shorter distance of 383 meters, and achieved a 97.96\% exploration rate. Even in the challenging environment of map M3, it achieved an exploration rate of 97.15\%. Overall, our algorithm outperformed the MRF approach in terms of map completeness, while time and distance traveled remained comparable to the MRF in the challenging environments of maps M3 and M4, demonstrating its ability to maintain an effective balance between thoroughness and efficiency.000
\begin{figure}[t]
  \centering
  \includegraphics[width=1\columnwidth]{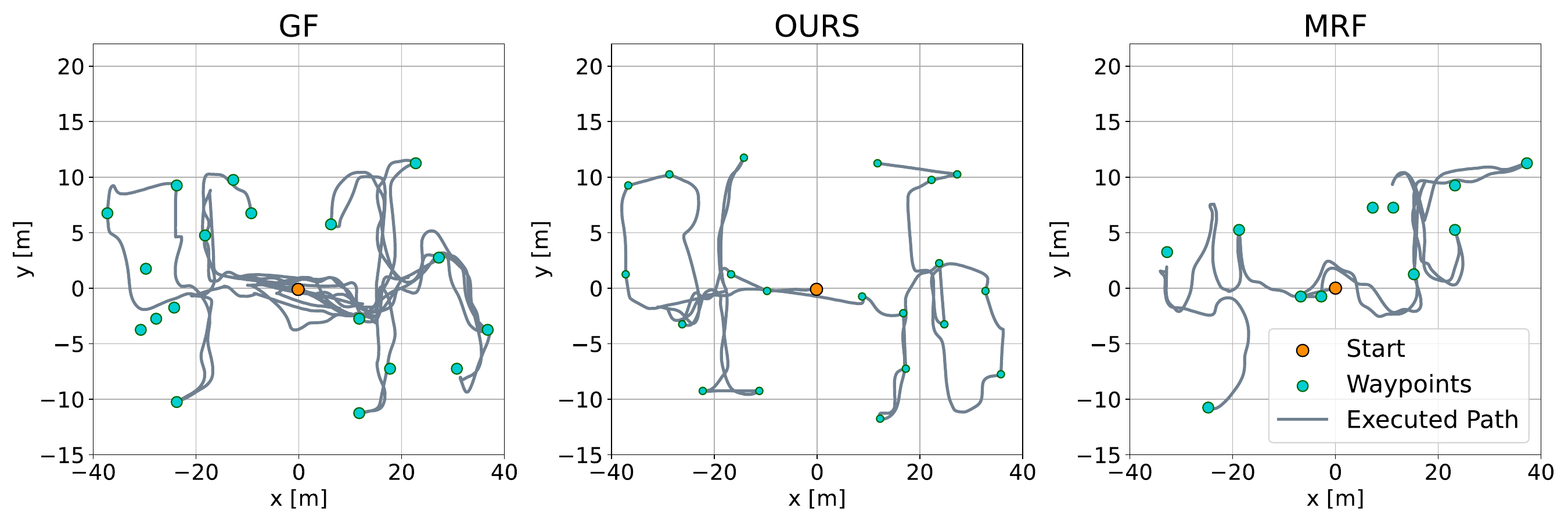}
  \caption{Paths during exploration of map 4 by all three approaches.}
  \label{fig:map4-trajectories}
  \vspace{10pt}
\end{figure}
\begin{figure}[t]
  \centering
  \includegraphics[width=0.9\columnwidth]{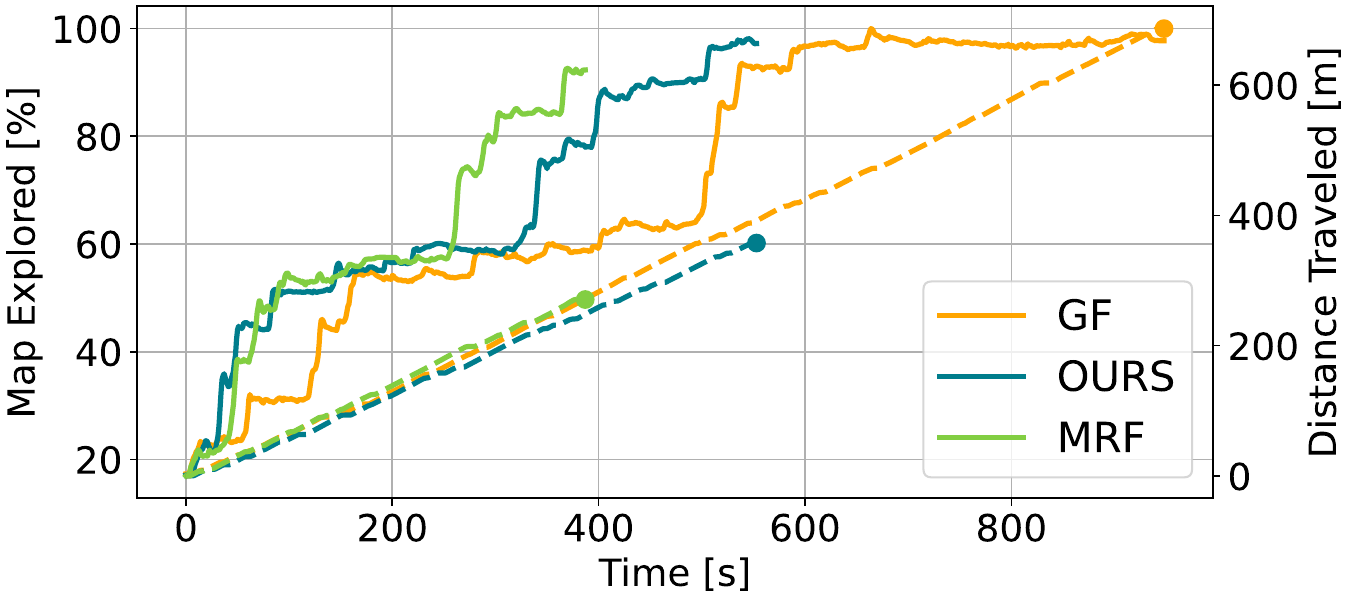}
  \caption{Percentage of explored volume (non-dashed) and distance traveled (dashed) comparison on map 4.}
  \vspace{-5pt}
  \label{fig:map4-results}
\end{figure}
The performance of the algorithms on map 4 is shown in Fig. \ref{fig:map4-trajectories} and Fig. \ref{fig:map4-results}. A clear illustration of the distinguishing features of the evaluated approaches is shown in Fig. \ref{fig:map4-trajectories}. The GF strategy is characterized by back-and-forth movements, while the MRF strategy neglects to visit significant portions of the map. Conversely, our strategy strikes a better balance between these extremes by having relatively fewer redundant paths and visiting more exploration targets. 


\vspace{-5pt}
\subsection{Outdoor Environments}
\label{subsec:outdoor}

\begin{figure}[t]
  \centering
  \begin{minipage}[b]{0.45\columnwidth}
    \centering
    \includegraphics[width=\columnwidth]{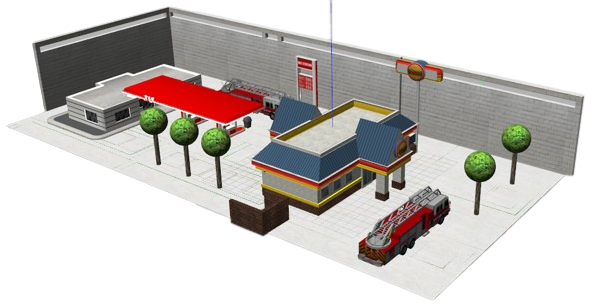}
    \subcaption{}
    \label{fig:Outdoors}
  \end{minipage}
  \hfill
  \begin{minipage}[b]{0.45\columnwidth}
    \centering
    \includegraphics[width=\columnwidth]{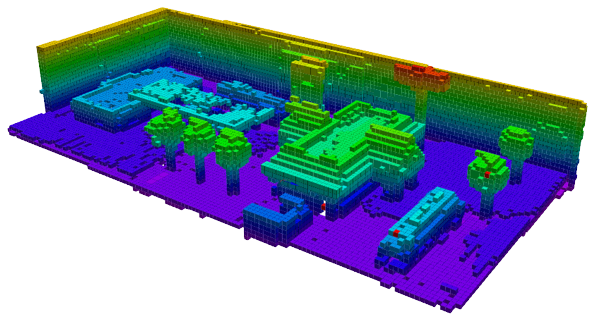}
    \subcaption{}
    \label{fig:resulting_octo}
  \end{minipage}
  \vspace{12pt}
  \caption{Exploration in an outdoor environment: (a) Gazebo world. (b) The OctoMap created during exploration.}
  \label{fig:3d-map}
  \vspace{10pt}
\end{figure}

\begin{figure}[t]
  \centering
  \includegraphics[width=0.9\columnwidth]{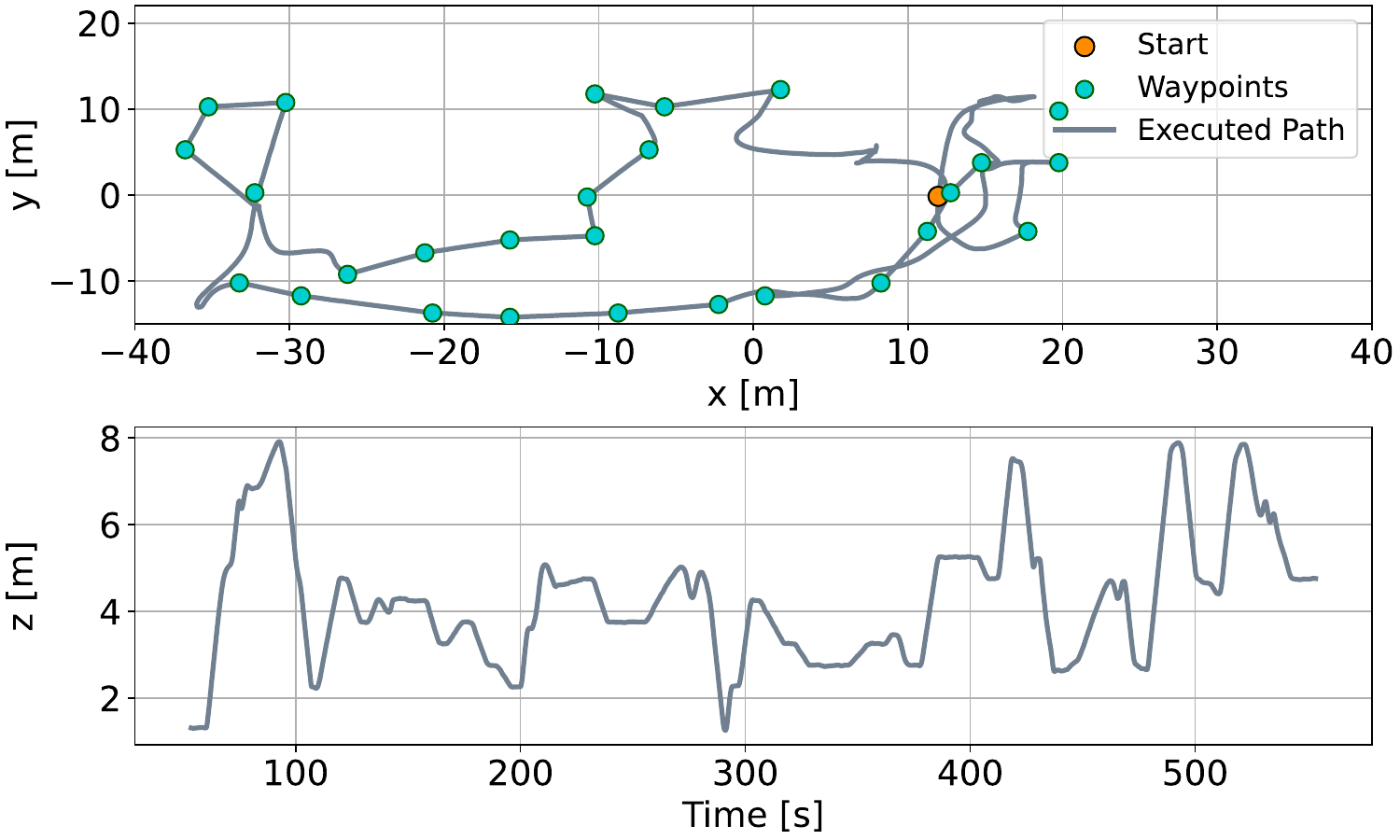}
  \caption{Paths during exploration of the outdoor environment.}
  \vspace{-5pt}
  \label{fig:map3d_trajectories}
\end{figure}

The scenario depicted in Fig. \ref{fig:3d-map} was specifically designed to examine and leverage the 3D capabilities of the proposed approach. The results of implementing our strategy are shown in Fig. \ref{fig:map3d_trajectories}, presenting the executed path in the x-y space and the flight altitude. It is noteworthy that there is minimal overlap between the exploration paths. Additionally, the elevation profile demonstrates how exploration targets are set at varied altitudes.

\section{Experimental evaluation}
\label{sec:experiment}
The proposed approach is also validated in a real-world environment. These tests aimed to verify the proper functioning and integration of the developed exploration solution within the UAV system. 
For our real-world experimental analysis, a \textit{HOLYBRO X500} quadcopter is used (Fig. \ref{fig:real_world} (a)). The dimensions of the UAV are 0.50 m $\times$ 0.50 m $\times$ 0.25 m, which makes it suitable for indoor environments. The total flight time of the UAV is around 18 min. Furthermore, the UAV is equipped with an \textit{Ouster 128-U} LiDAR with a maximum range of 200 m. The maximum range is reduced to 5 m to fit the exploration area of 15 m $\times$ 15 m $\times$ 5 m (Fig. \ref{fig:real_world} (b)).
All the experiments have been carried out in an indoor testbed, which houses a VICON localization system, that can easily accommodate structures to recreate custom scenarios for a wide range of aerial robotic applications.
Experimental evaluations were tested using $r = 0.25$ m.

\begin{figure}[t]
  \centering
  \begin{minipage}{0.5\columnwidth}
    \centering
    \includegraphics[width=\columnwidth]{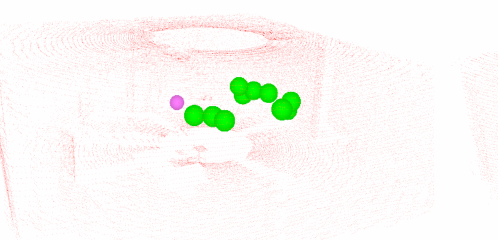}
    \subcaption{}
    \label{fig:rviz_candidates}
  \end{minipage}
  \hfill
  \begin{minipage}{0.47\columnwidth}
    \centering
    \includegraphics[width=\columnwidth]{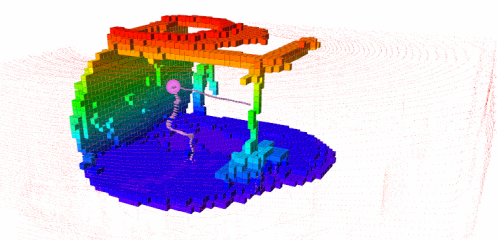}
    \subcaption{}
    \label{fig:rviz_path}
  \end{minipage}
  \vspace{12pt}
  \caption{(a) Exploration candidates (green) and the selected goal (pink) in a single iteration. (b) Generated OctoMap of the environment and the executed path.}
  \label{fig:real_world_rviz}
  \vspace{10pt}
\end{figure}

The result of the exploration is the OctoMap of the environment shown in Fig. \ref{fig:real_world_rviz}.
Running the proposed exploration strategy in the real world and in real time, the successful exploration and mapping are demonstrated while running the localization, OctoMap creation, exploration strategy and control algorithms on the UAV with limited onboard resources.

\begin{figure}[t]
  \centering
    \includegraphics[width=0.9\columnwidth]{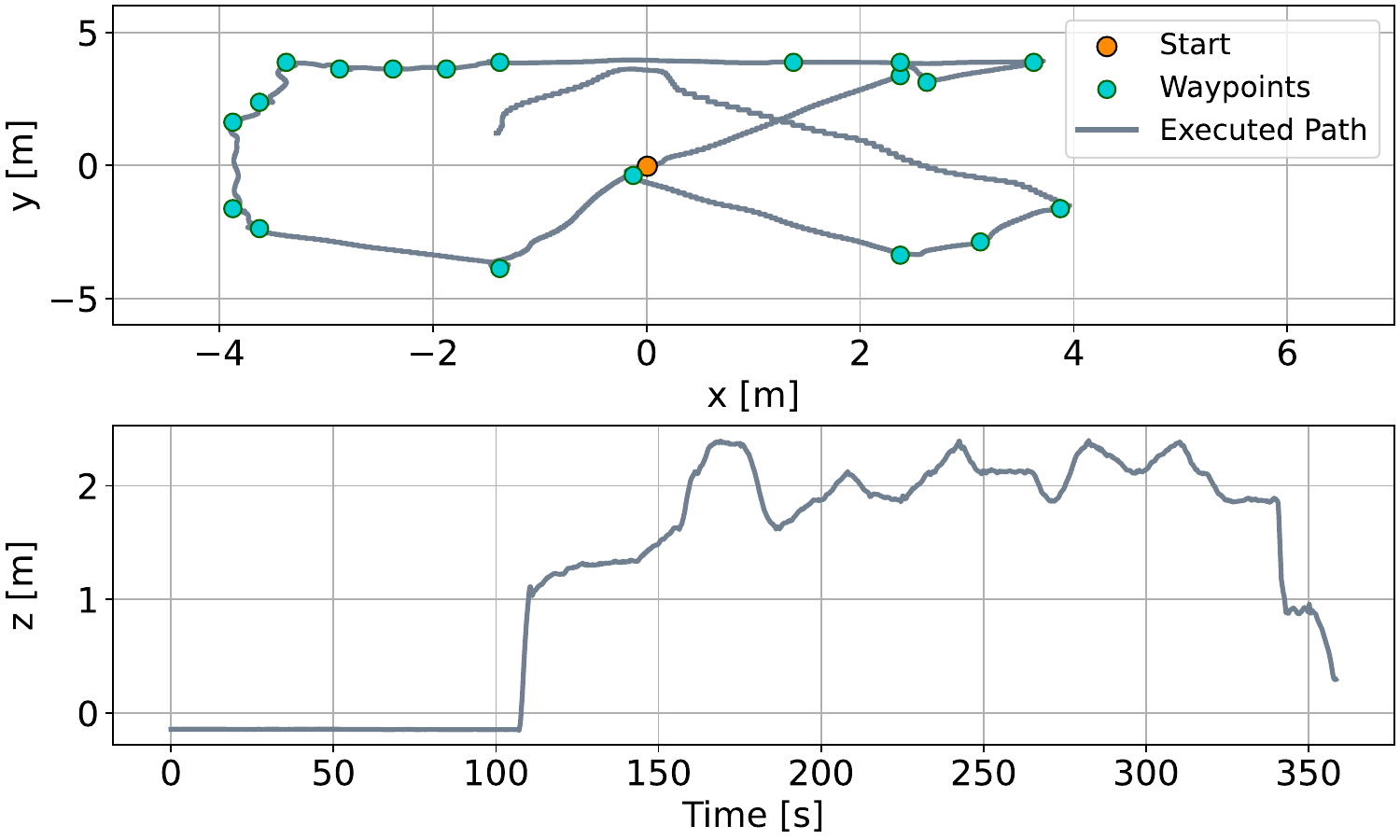}
    \caption{Paths during exploration of the real-world environment.}
    \vspace{-5pt}
  \label{fig:testbed-trajectories}
\end{figure}

The behavior and performance of the proposed exploration strategy are displayed in Fig. \ref{fig:testbed-trajectories}, showing the executed path in the x-y space and the visited points during the exploration, as well as the altitude profile of the UAV during the flight.
Even though these were short flights, they were incredibly valuable for verifying the proper functioning of the entire exploration loop and the developed strategy, including frontier detection, selection, path planning, and navigation.  It is worth highlighting that all algorithms were running onboard the drone, which shows that the computational effort does not exceed the limits of the available onboard computer of small UAVs.

\section{Conclusion}
\label{sec:conclusion}

This paper proposes a novel frontier-based exploration strategy that efficiently balances exploration speed and information acquisition. The proposed method is validated through both simulated and real-world tests, showing improved behavior in terms of the time required to completely explore and create a map of the environment compared to state-of-the-art strategies. While each strategy has its merits, the proposed algorithm stands out for its balance of speed and thoroughness in complex environments. 

Video recordings of the proposed 3D exploration strategy can be found on YouTube \cite{video-exploration}.

\appendices
\section*{Acknowledgements}
\footnotesize{This work has been supported in part by the scientific project Strengthening Research and Innovation Excellence in Autonomous Aerial Systems - AeroSTREAM supported by European Commission HORIZON-WIDERA-2021-ACCESS-05 Programme through project under G. A. number 101071270, and BEEYONDERS project (Horizon programme grant agreement No 101058548).
The work of doctoral student Ana Milas has been supported in part by the “Young researchers’ career development project--training of doctoral students” of the Croatian Science Foundation funded by the European Union from the European Social Fund.}

\bibliographystyle{ieeetr}
\typeout{}
\balance
\bibliography{Bibliography/exploration}

\begin{thebibliography}{10}

\bibitem{CamposMacas2020}
L.~Campos-Mac{\'{\i}}as, R.~Aldana-L{\'{o}}pez, R.~Guardia, J.~I.
  Parra-Vilchis, and D.~G{\'{o}}mez-Guti{\'{e}}rrez, ``Autonomous navigation of
  {MAVs} in unknown cluttered environments,'' {\em Journal of Field Robotics},
  vol.~38, pp.~307--326, May 2020.

\bibitem{Kwon2020}
W.~Kwon, J.~H. Park, M.~Lee, J.~Her, S.-H. Kim, and J.-W. Seo, ``Robust
  autonomous navigation of unmanned aerial vehicles (uavs) for warehouses’
  inventory application,'' {\em IEEE Robotics and Automation Letters}, vol.~5,
  no.~1, pp.~243--249, 2020.

\bibitem{Yamauchi1997}
B.~{Yamauchi}, ``A frontier-based approach for autonomous exploration,'' in
  {\em Proceedings 1997 IEEE International Symposium on Computational
  Intelligence in Robotics and Automation CIRA'97.}, pp.~146--151, 1997.

\bibitem{Bircher2016}
A.~{Bircher}, M.~{Kamel}, K.~{Alexis}, H.~{Oleynikova}, and R.~{Siegwart},
  ``Receding horizon "next-best-view" planner for 3{D} exploration,'' in {\em
  2016 IEEE International Conference on Robotics and Automation (ICRA)},
  pp.~1462--1468, 2016.

\bibitem{Cieslewski2017}
T.~Cieslewski, E.~Kaufmann, and D.~Scaramuzza, ``Rapid exploration with
  multi-rotors: A frontier selection method for high speed flight,'' in {\em
  2017 {IEEE}/{RSJ} International Conference on Intelligent Robots and Systems
  ({IROS})}, 2017.

\bibitem{Witting2018}
C.~Witting, M.~Fehr, R.~Bähnemann, H.~Oleynikova, and R.~Siegwart,
  ``History-aware autonomous exploration in confined environments using
  {MAV}s,'' in {\em 2018 IEEE/RSJ International Conference on Intelligent
  Robots and Systems (IROS)}, pp.~1--9, 2018.

\bibitem{Dai2020}
A.~Dai, S.~Papatheodorou, N.~Funk, D.~Tzoumanikas, and S.~Leutenegger, ``Fast
  frontier-based information-driven autonomous exploration with an {MAV},'' in
  {\em 2020 {IEEE} International Conference on Robotics and Automation
  ({ICRA})}, pp.~9570--9576, 2020.

\bibitem{Selin2019}
M.~Selin, M.~Tiger, D.~Duberg, F.~Heintz, and P.~Jensfelt, ``Efficient
  autonomous exploration planning of large-scale 3{D} environments,'' {\em IEEE
  Robotics and Automation Letters}, vol.~4, no.~2, pp.~1699--1706, 2019.

\bibitem{Batinovic-RAL-2021}
A.~Batinovic, T.~Petrovic, A.~Ivanovic, F.~Petric, and S.~Bogdan, ``A
  multi-resolution frontier-based planner for autonomous 3{D} exploration,''
  {\em IEEE Robotics and Automation Letters}, vol.~6, no.~3, pp.~4528--4535,
  2021.

\bibitem{Batinovic-RAL-2022}
A.~Batinovic, A.~Ivanovic, T.~Petrovic, and S.~Bogdan, ``A shadowcasting-based
  next-best-view planner for autonomous 3d exploration,'' {\em IEEE Robotics
  and Automation Letters}, vol.~7, no.~2, pp.~2969--2976, 2022.

\bibitem{RRT1}
S.~M. Lavalle, J.~J. Kuffner, and Jr., ``Rapidly-exploring random trees:
  Progress and prospects,'' in {\em Algorithmic and Computational Robotics: New
  Directions}, pp.~293--308, 2000.

\bibitem{github-collector}
``{A}utonomous {E}xploration of {U}nknown {3D} {E}nvironments {U}sing a
  {F}rontier-{B}ased {C}ollector {S}trategy: {O}pen {S}ource.''
  \url{https://github.com/davidchangoluisa/uav\_frontier\_based\_collector}.

\bibitem{Zhu2015}
C.~{Zhu}, R.~{Ding}, M.~{Lin}, and Y.~{Wu}, ``A 3{D} frontier-based exploration
  tool for mavs,'' in {\em 2015 IEEE 27th International Conference on Tools
  with Artificial Intelligence (ICTAI)}, pp.~348--352, 2015.

\bibitem{Faria2019}
M.~Faria, R.~Mar{\'{\i}}n, M.~Popovi{\'{c}}, I.~Maza, and A.~Viguria,
  ``Efficient lazy theta* path planning over a sparse grid to explore large
  3{D} volumes with a multirotor {UAV},'' {\em Sensors}, vol.~19, no.~1,
  p.~174, 2019.

\bibitem{zhou2020survey}
X.~Zhou, Z.~Yi, Y.~Liu, K.~Huang, and H.~Huang, ``Survey on path and view
  planning for uavs,'' {\em Virtual Reality \& Intelligent Hardware}, vol.~2,
  no.~1, pp.~56--69, 2020.

\bibitem{Schmid2020}
L.~Schmid, M.~Pantic, R.~Khanna, L.~Ott, R.~Siegwart, and J.~Nieto, ``An
  efficient sampling-based method for online informative path planning in
  unknown environments,'' {\em IEEE Robotics and Automation Letters}, vol.~5,
  no.~2, pp.~1500--1507, 2020.

\bibitem{Respall2021}
V.~{Massague Respall}, D.~{Devitt}, R.~{Fedorenko}, and A.~{Klimchik}, ``Fast
  sampling-based next-best-view exploration algorithm for a {MAV},'' in {\em
  2021 {IEEE} International Conference on Robotics and Automation ({ICRA})},
  2021.

\bibitem{zhou2021fuel}
B.~Zhou, Y.~Zhang, X.~Chen, and S.~Shen, ``Fuel: Fast uav exploration using
  incremental frontier structure and hierarchical planning,'' {\em IEEE
  Robotics and Automation Letters}, vol.~6, no.~2, pp.~779--786, 2021.

\bibitem{zhou2023racer}
B.~Zhou, H.~Xu, and S.~Shen, ``Racer: Rapid collaborative exploration with a
  decentralized multi-uav system,'' {\em IEEE Transactions on Robotics}, 2023.

\bibitem{Oleynikova2018}
H.~Oleynikova, Z.~Taylor, R.~Siegwart, and J.~Nieto, ``Safe local exploration
  for replanning in cluttered unknown environments for microaerial vehicles,''
  {\em IEEE Robotics and Automation Letters}, vol.~3, no.~3, pp.~1474--1481,
  2018.

\bibitem{Zhao2023}
Y.~Zhao, L.~Yan, H.~Xie, J.~Dai, and P.~Wei, ``Autonomous exploration method
  for fast unknown environment mapping by using uav equipped with limited fov
  sensor,'' {\em IEEE Transactions on Industrial Electronics}, pp.~1--10, 2023.

\bibitem{Hornung2013}
A.~Hornung, K.~M. Wurm, M.~Bennewitz, C.~Stachniss, and W.~Burgard,
  ``{OctoMap}: an efficient probabilistic 3{D} mapping framework based on
  octrees,'' {\em Autonomous Robots}, vol.~34, no.~3, pp.~189--206, 2013.

\bibitem{Fukunaga1975}
K.~Fukunaga and L.~Hostetler, ``The estimation of the gradient of a density
  function, with applications in pattern recognition,'' {\em {IEEE}
  Transactions on Information Theory}, vol.~21, no.~1, pp.~32--40, 1975.

\bibitem{Arbanas2018}
B.~Arbanas, A.~Ivanovic, M.~Car, M.~Orsag, T.~Petrovic, and S.~Bogdan,
  ``Decentralized planning and control for {UAV}-{UGV} cooperative teams,''
  {\em Autonomous Robots}, vol.~42, no.~8, pp.~1601--1618, 2018.

\bibitem{Ivanovic2022ParabolicAirdrop}
A.~Ivanovic and M.~Orsag, ``Parabolic airdrop trajectory planning for
  multirotor unmanned aerial vehicles,'' {\em IEEE Access}, vol.~10,
  pp.~36907--36923, 2022.

\bibitem{larics-motion-planning}
UNIZG-FER and LARICS, ``Path and trajectory planning.''
  https://github.com/larics/larics\_motion\_planning.
\newblock Accessed: 2023-06-06.

\bibitem{tomas_mpc}
T.~Baca, D.~Hert, G.~Loianno, M.~Saska, and V.~Kumar, ``Model predictive
  trajectory tracking and collision avoidance for reliable outdoor deployment
  of unmanned aerial vehicles,'' in {\em 2018 IEEE/RSJ International Conference
  on Intelligent Robots and Systems (IROS)}, pp.~6753--6760, 2018.

\bibitem{video-exploration}
``{A}utonomous {E}xploration of {U}nknown {3D} {E}nvironments {U}sing a
  {F}rontier-{B}ased {C}ollector {S}trategy: {V}ideo.''
  \url{https://www.youtube.com/watch?v=f5a5eGt-Hqs}.

\end{thebibliography}

\end{document}